%% file: main.tex
\documentclass[runningheads]{llncs}
\usepackage[T1]{fontenc}
\usepackage[utf8]{inputenc} 
\usepackage{url}            
\usepackage{booktabs}       
\usepackage{amsfonts}       
\usepackage{nicefrac}       
\usepackage{microtype}      
\usepackage[dvipsnames,table]{xcolor}         
\usepackage{subcaption}
\usepackage[ruled]{algorithm2e}
\SetAlFnt{\small}
\usepackage{amsmath}
\usepackage{bm}
\usepackage{bbold}
\usepackage{graphicx}
\usepackage{multirow}
\usepackage{textcomp}
\usepackage{wrapfig}
\usepackage{float}
\usepackage{hyperref} 

\definecolor{lightblue}{rgb}{0.0, 0.2, 0.9}

\hypersetup{
    colorlinks=true,
    linkcolor=lightblue,
    filecolor=lightblue,      
    urlcolor=lightblue,
    citecolor=lightblue,
}
\usepackage{subcaption}

\usepackage{xspace}
\usepackage{interval}
\usepackage{bm,upgreek}
\usepackage{colortbl}
\usepackage{enumitem}
\usepackage{amssymb}
\usepackage{pifont}
\usepackage{graphicx}
\usepackage{color}
\usepackage{multirow}
\usepackage{amsmath,amssymb}
\usepackage{mathtools,leftindex,tensor,mhchem}
\usepackage[table]{xcolor}
\usepackage[misc]{ifsym}

\makeatletter
\newcommand{\smalloplus}{\mathbin{\mathpalette\make@small\oplus}}
\newcommand{\smallotimes}{\mathbin{\mathpalette\make@small\otimes}}
\newcommand{\make@small}[2]{%
  \vcenter{\hbox{%
    \scalebox{0.8}{$\m@th#1#2$}%
  }}%
}
\makeatother
\definecolor{darkgreen}{RGB}{0,150,0}
\usepackage{pifont}
\newcommand{\cmark}{\textcolor{darkgreen}{\ding{51}}}
\newcommand{\xmark}{\textcolor{red}{\ding{55}}}

\definecolor{greencheck}{RGB}{0,128,0}
\definecolor{redcross}{RGB}{255,0,0}

\DeclareMathOperator*{\argmax}{arg\,max}
\DeclareMathOperator*{\argmin}{arg\,min}

\begin{document}
\title{PromptSmooth: Certifying Robustness of Medical Vision-Language Models via Prompt Learning}
\titlerunning{PromptSmooth}
\author{Noor Hussein\inst{(\textrm{\Letter})} \and Fahad Shamshad \and Muzammal Naseer \and Karthik Nandakumar}

\authorrunning{N. Hussein et al.}
\institute{Mohamed Bin Zayed University of Artificial Intelligence, Abu Dhabi, UAE \\
\email{\{noor.hussein, fahad.shamshad, muzammal.naseer, karthik.nandakumar\}@mbzuai.ac.ae}}

\maketitle              
\renewcommand{\thefootnote}{}
\footnotetext{\inst{\textrm{\Letter}}Corresponding Author}
\renewcommand{\thefootnote}{\arabic{footnote}}
\setcounter{footnote}{0}

\begin{abstract}
Medical vision-language models (Med-VLMs) trained on large datasets of medical image-text pairs and later fine-tuned for specific tasks have emerged as a mainstream paradigm in medical image analysis. However, recent studies have highlighted the susceptibility of these Med-VLMs to adversarial attacks, raising concerns about their safety and robustness. Randomized smoothing is a well-known technique for turning any classifier into a model that is certifiably robust to adversarial perturbations. However, this approach requires retraining the Med-VLM-based classifier so that it classifies well under Gaussian noise, which is often infeasible in practice. In this paper, we propose a novel framework called \texttt{PromptSmooth} to achieve efficient certified robustness of Med-VLMs by leveraging the concept of prompt learning. Given any pre-trained Med-VLM, \texttt{PromptSmooth} adapts it to handle Gaussian noise by learning textual prompts in a zero-shot or few-shot manner, achieving a delicate balance between accuracy and robustness, while minimizing the computational overhead. Moreover, \texttt{PromptSmooth} requires only a single model to handle multiple noise levels, which substantially reduces the computational cost compared to traditional methods that rely on training a separate model for each noise level. Comprehensive experiments based on three Med-VLMs and across six downstream datasets of various imaging modalities demonstrate the efficacy of \texttt{PromptSmooth}. Our code and models are available at \href{https://github.com/nhussein/promptsmooth}{\color{Magenta}{{https://github.com/nhussein/promptsmooth}}}.

\end{abstract}

\section{Introduction} \label{sec:intro}
Medical Vision-Language Models (Med-VLMs) have significantly advanced the state-of-the-art across a broad spectrum of medical imaging  tasks such as classification, segmentation, and detection~\cite{shrestha2023medical,zhao2023clip}. During pre-training, these models learn generic representations from large volumes of medical image-text pairs and subsequently transfer this knowledge to downstream medical tasks, which often suffer from limited data availability~\cite{zhang2023prompt}. However, recent advances in adversarial machine learning have exposed the vulnerability of VLMs to adversarial attacks~\cite{zhao2024evaluating}, which introduce small,  imperceptible perturbations to the image that drastically change the resulting predictions. Med-VLMs are also prone to these attacks~\cite{finlayson2019adversarial,han2023medical}, which poses a significant risk to the integrity of medical diagnostics, underscoring the need for defense mechanisms to safeguard against such threats.

\noindent Though many empirical approaches have been proposed to defend medical models against adversarial attacks~\cite{dong2023adversarial}, these defenses have consistently shown vulnerabilities to newer and more powerful adversarial attacks~\cite{athalye2018obfuscated}. Consequently, \textit{certifiable defenses}~\cite{li2023sok} with provable adversarial robustness guarantees have attracted considerable attention, particularly in the safety-critical medical domain~\cite{laousy2023certification}. Specifically, these \textit{certifiable defenses} guarantee that the model’s predictions will remain unchanged for adversarial perturbations bounded by a \textit{certified radius} around an input sample. However, most of these \textit{certified defenses} are either not scalable to large models or have been evaluated on low-dimensional datasets (\textit{e.g.}, $32\times32$)~\cite{kumari2023trust}, significantly hindering their applicability to Med-VLMs and/or high-dimensional datasets encountered in medical imaging~\cite{azad2023foundational}.

\begin{table}[t]
\centering
\setlength{\tabcolsep}{2.5pt}
\caption{Comparison of different randomized smoothing implementations.}
\resizebox{\linewidth}{!}{
\begin{tabular}{lccccc}
\toprule
\rowcolor{gray!20} \textbf{Methods} & \begin{tabular}[c]{@{}c@{}}{Data}\\ {Efficient}\end{tabular} & {\begin{tabular}[c]{@{}c@{}}{Computational}\\ {Cost}\end{tabular}} & {\begin{tabular}[c]{@{}c@{}}{Noise-Agnostic}\\ {Training}\end{tabular}} & {\begin{tabular}[c]{@{}c@{}}{Tailored}\\ {to VLM}\end{tabular}} & {\begin{tabular}[c]{@{}c@{}}{Accuracy vs.}\\ {Robustness Trade-off}\end{tabular}} \\ \midrule
{\color{orange}{Noise-augmented Re-training}}~\cite{cohen2019certified} & \Large{\xmark} & \textcolor{red}{\textbf{High}} & \Large{\xmark} & \Large{\xmark} &  \textcolor{red}{\textbf{High}} \\
{\color{ProcessBlue}{Denoised Smoothing}}~\cite{salman2020denoised} & \Large{\xmark}  &  \textcolor{red}{\textbf{High}} & \Large{\xmark} & \Large{\xmark} & Moderate\\
{\color{RubineRed}{Diffusion Smoothing}}~\cite{carlini2022certified} & \Large{\xmark} & Moderate & \Large{\cmark} & \Large{\xmark} & \textcolor{OliveGreen}{\textbf{Low}} \\ \midrule
\rowcolor{Apricot!20}  \textbf{\texttt{PromptSmooth} (Ours)} & \Large{\cmark} & \textcolor{OliveGreen}{\textbf{Low}} & \Large{\cmark} & \Large{\cmark} & \textcolor{OliveGreen}{\textbf{Low}} \\ \bottomrule
\end{tabular}
}
\vspace{-1.5em}
\label{tab:challenges_medical_vlms}
\vspace{0em}
\end{table}

\noindent A well-known approach for addressing the scalability issue is randomized smoothing (RS)~\cite{lecuyer2019certified}, which constructs a new \textit{smoothed classifier} by averaging the output of a \textit{base classifier} under random Gaussian perturbations of the input. The addition of Gaussian noise to the input image creates a trade-off between accuracy and robustness~\cite{li2023sok}, which depends on how well the base classifier performs on noisy images. As the noise variance increases, robustness improves at the cost of lower clean accuracy. 
To improve the trade-off between accuracy and robustness, three broad strategies have been proposed.
The {\color{orange}{\textit{first}}} approach involves training a classifier from scratch on a Gaussian noise-augmented dataset~\cite{cohen2019certified,qiu2023exploring}.
The {\color{ProcessBlue}{\textit{second}}} strategy prepends a custom-trained denoiser before the pre-trained classifier to remove Gaussian noise from the image prior before RS~\cite{salman2020denoised}.
The {\color{RubineRed}{\textit{third}}} approach utilizes pre-trained off-the-shelf diffusion models (trained on large-scale image datasets) as denoisers~\cite{carlini2022certified,laousy2023certification}. 
Extending these methods to Med-VLMs presents unique challenges (see Tab.~\ref{tab:challenges_medical_vlms}). {\color{orange}{\textit{Noise-augmented re-training}}} of Med-VLMs would require substantial computational resources and access to large (often privacy-sensitive) medical datasets.
{\color{ProcessBlue}{\textit{Denoiser prepending}}} requires a large dataset of paired clean-noisy images as well as time-consuming denoiser training for each noise level. {\color{RubineRed}{\textit{Diffusion-based denoisers}}} require extensive datasets to accurately model complex medical images and training of such diffusion models is expensive.

\noindent To overcome the above limitations, we propose \texttt{PromptSmooth} to efficiently achieve certified robustness in pre-trained Med-VLMs without hampering clean accuracy. Instead of re-training the VLM from scratch or utilizing denoisers, we inject a small number of learnable prompts (tokens) into the VLM input space and optimize them, while keeping the entire backbone frozen. Our contributions are two-fold: (i) To the best of our knowledge, this is the first work where prompt learning is exploited for efficient robustness certification of Med-VLMs in classification, and (ii) We propose algorithms for effective prompt learning under both zero-shot (\texttt{Zero-Shot PromptSmooth}) and few-shot (\texttt{Few-Shot PromptSmooth}) settings.

\section{Related Work and Background} \label{sec:related_work}
\textbf{Medical VLMs}: 
Medical VLMs based on Contrastive Language Image Pre-training (CLIP)~\cite{radford2021learning} have gained considerable attention in medical imaging~\cite{zhao2023clip}. This pre-training method aims to maximize the cosine similarity between the embeddings of matched image-text pairs, while minimizing it among unmatched pairs.
Despite the introduction of numerous Med-VLMs for many imaging modalities, including histopathology~\cite{ikezogwo2024quilt,huang2023visual}, X-ray~\cite{wang2022medclip}, and retinal~\cite{silva2023foundation} images, a critical evaluation of their robustness remains largely unexplored.

\noindent \textbf{Certified Robustness}: Let $f:\mathcal{X} \rightarrow \mathcal{Y}$ be a base classifier that maps an input $\mathbf{x} \in \mathcal{X} \subseteq \mathbb{R}^D$ into a class label $y \in \mathcal{Y} = \{1,2,\cdots,K\}$, where $\mathcal{X}$ and $\mathcal{Y}$ are the input and label spaces, $D$ is the input dimensionality, and $K$ is the number of classes. Randomized smoothing (RS) \cite{cohen2019certified} transforms the base classifier $f$ into a smoothed classifier $g$ as follows: $g(\mathbf{x}) = \argmax_{y \in \mathcal{Y}} \; \mathbb{P}[f(\mathbf{x}+\delta) = y]$, where $\delta \sim \mathcal{N}(0, \sigma^2 \mathbf{I})$ and $\mathbb{P}$ denotes a probability measure. For an input $\mathbf{x}$, $g$ predicts the class most likely under the base classifier $f$ when $\mathbf{x}$ is perturbed with isotropic Gaussian noise $\delta$. 
RS provides the following robustness guarantee for $g$: if $y_A \in \mathcal{Y}$ and $\underline{p_A},\overline{p_B} \in [0,1]$ satisfy $\mathbb{P}[f(\mathbf{x}+\delta) = y_A] \geq \underline{p_A} \geq \overline{p_B} \geq \max_{y \neq y_A} \mathbb{P}[f(\mathbf{x}+\delta) = y]$, then $g(\mathbf{x}+\mathbf{r})=y_A$ for all $||\mathbf{r}||_2 < R$, where the certified radius $R$ around an input $\mathbf{x}$ is given by $R = \frac{\sigma}{2} \left(\Phi^{-1}(\underline{p_A}) - \Phi^{-1}(\overline{p_B})\right)$. This guarantee ensures that for any $\ell_2$ adversarial perturbation $\mathbf{r}$ with magnitude less than $R$, the output of the smoothed classifier $g$ remains unchanged. Here, $\underline{p_A}$ and $\overline{p_B}$ are the lower-bound and upper-bound of probabilities of the most-likely ($y_A$) and second-most-likely ($y_B$) classes, respectively, predicted by $f$ under noise, and $\Phi^{-1}$ is the inverse of the standard Gaussian cdf. For practical applications, RS uses Monte Carlo sampling to estimate $\underline{p_A}$ and $\overline{p_B}$, thereby facilitating the computation of a certified radius.
Increasing noise variance $\sigma$ leads to better robustness (higher $R$), but at the cost of accuracy (because predictions of $f$ under noise become less reliable). This trade-off can be mitigated by improving the accuracy of $f$ under noise. 

\noindent \textbf{Prompt Learning:} Prompt learning (PL) is a technique that fine-tunes VLMs for specific tasks by adding learnable prompt tokens to the model's input, thereby avoiding changes to existing parameters. 
The effectiveness of PL in few-shot scenarios~\cite{zhou2022learning,zhong2023ariadne} makes it especially useful for data-limited medical imaging tasks. Recently, attempts have also been made to learn prompts in a zero-shot manner by enforcing consistency regularization between multiple augmentations of a test sample at test time~\cite{shu2022test}. While PL is typically used to improve performance on downstream tasks,  this work investigates how to leverage PL for efficient robustness certification of Med-VLMs in both few-shot and zero-shot settings.

\section{Methodology} \label{sec:methodology}

Our \textit{goal} is to efficiently adapt zero-shot classifiers based on Med-VLMs in data-limited scenarios to predict well under Gaussian noise, thereby ensuring that they maintain high accuracy on clean images while also achieving strong certified robustness. 
We first outline how Med-VLMs can be used for zero-shot inference on downstream tasks and introduce \texttt{PromptSmooth} for their efficient adaptation in few/zero-shot settings.

\begin{figure}[t]
  \centering
  \includegraphics[width=\textwidth]{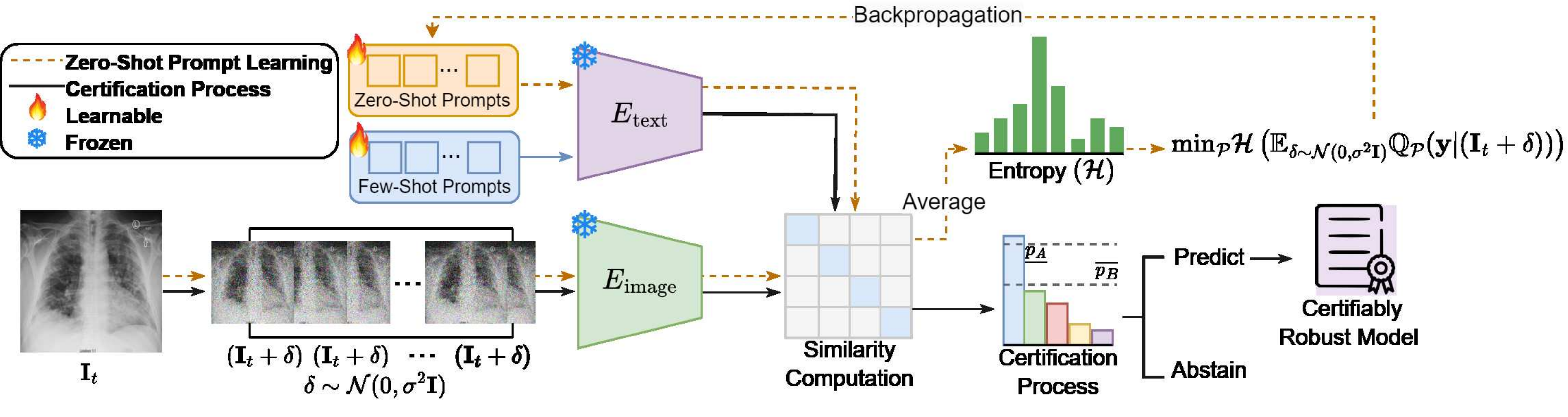}
  \caption{\small Overview of \texttt{PromptSmooth} for certified robustness. Prompts can be learned {\color{ProcessBlue}{offline}}  or {\color{orange}{at test-time}}. Gaussian noise is added at test-time to $T$ copies of the input $\mathbf{I_t}$ and prompts are learned by minimizing the entropy loss ({\color{orange}{dashed orange line}}). Using {\color{orange}{zero-shot}} and/or {\color{ProcessBlue}{few-shot}} prompts, inference is repeated for $M$ noisy instances for certification (solid black line). Model predicts (and gives a certified radius) or abstains.}
  \label{fig:promptmsooth_method}
\end{figure}

\subsection{Zero-shot Inference based on Med-VLMs} \label{sec:zeroshot}

Med-VLMs learn an alignment between image and text input spaces (denoted as $\mathcal{I}$ and $\mathcal{T}$, respectively) and typically consist of two encoders: an image encoder $\mathbf{E}_{\text{image}}: \mathcal{I} \rightarrow \mathbb{R}^d$ and a text encoder $\mathbf{E}_{\text{text}}: \mathcal{T} \rightarrow \mathbb{R}^d$. The image encoder maps a given image $\mathbf{I} \in \mathcal{I} \subseteq \mathbb{R}^{H \times W \times C}$ 
into a $d$-dimensional image feature vector $\mathbf{v} \in  \mathbb{R}^d$. Similarly, the text encoder maps the given text $\mathbf{T} \in \mathcal{T}$ into a text feature vector $\mathbf{u} \in \mathbb{R}^d$. These models utilize a contrastive loss during pre-training to enhance the similarity between text and image feature vectors, ensuring their alignment within the feature space.
After pre-training, Med-VLMs can be used in the zero-shot manner for various downstream tasks like image classification.
For zero-shot application, consider a test image $\mathbf{I_{t}} \in \mathcal{I}$ from class $y_{t} \in \mathcal{Y}$. All the class labels $y_i \in \mathcal{Y}$ ($i \in [1,K]$) are converted into text prompts using a hand-crafted template such as $\mathbf{t}(y_i) = ``\texttt{A X-ray image of [CLASS} ~y_i\texttt{] patient}''$. These text prompts are processed by the text encoder to obtain $\{\mathbf{u}_1, \mathbf{u}_2, \cdots, \mathbf{u}_K\}$, where $\mathbf{u}_i = \mathbf{E}_{\text{text}}(\mathbf{t}(y_i))$. Let $\mathbf{v}_{t} = \mathbf{E}_{\text{image}}(\mathbf{I}_{t})$ be the image feature vector for the test image. A cosine similarity score $ s_i = \text{sim}(\mathbf{u}_i, \mathbf{v}_{t})$ is computed for $i \in [1,K]$ and the prediction probabilities for $\mathbf{I}_{t}$ are obtained as $\mathbb{P}(y_i|\mathbf{I}_{t}) = \frac{\exp(\tau s_i)}{\sum_{j=1}^{K}\exp(\tau s_j)},$ where $\tau$ is the softmax temperature parameter. Thus, a zero-shot classifier $f$ based on the Med-VLM $(\mathbf{E}_{\text{image}},\mathbf{E}_{\text{text}})$ outputs a predicted label $\hat{y}_t$, where $\hat{y}_t = f(\mathbf{I}_{t}) = \argmax_{y \in \mathcal{Y}} \mathbb{P}(y|\mathbf{I}_{t})$. 
Despite their impressive zero-shot capabilities, Med-VLMs cannot be directly subjected to RS as they are pre-trained on clean datasets and their accuracy drops drastically when input images are perturbed with Gaussian noise. A naive solution is to pre-train the Med-VLMs from scratch with noisy data augmentation as in~\cite{cohen2019certified}, but this is often practically infeasible.

\subsection{PromptSmooth}
We now present our \texttt{PromptSmooth} approach, as shown in Figure~\ref{fig:promptmsooth_method}, to efficiently adapt a zero-shot classifier $f$ based on Med-VLMs such that high certified adversarial robustness can be achieved without severely degrading clean accuracy. The key idea is to inject small number of learnable prompts as inputs to the text encoder of the Med-VLM and learn these prompts to improve prediction accuracy on noisy images, while keeping the backbone fixed. Note that when a text prompt $\mathbf{t}(y_i) \in \mathcal{T}$ is presented as input to the text encoder, it is broken down into a sequence of word tokens, with their embeddings processed by the encoder. In other words, $\mathbf{t}(y_i)$ can be represented as a sequence $[\mathbf{w}]_1~[\mathbf{w}]_2~\cdots~[\texttt{CLASS}~y_i]$, where $[*]$ represents the embedding of a single word in the text prompt. In prompt learning (PL), the fixed word embeddings (except for the class name) are replaced with $M$ learnable embeddings, \textit{i.e.},  $\mathbf{t}(y_i)$ can now be represented as a sequence $\mathbf{p}_{i1}~\mathbf{p}_{i2}~\cdots\mathbf{p}_{iM}~[\texttt{CLASS}~y_i]$, where the dimensionality of $\mathbf{p}$ is the same as $[\mathbf{w}]$. Let $\mathcal{P} = \{\mathbf{p}_{im}\}$, $i \in [1,K]$, $m \in [1,M]$, denote the collection of all learnable prompts. Also, let $\mathbf{u}_i(\mathcal{P})$ be the text feature vector output by the text encoder for class $y_i$ after introduction of the learnable prompts $\mathcal{P}$ and $f_{\mathcal{P}}$ be the modified zero-shot classifier based on these new text features. Next, we address the question of how to learn these prompts $\mathcal{P}$ efficiently. \\
\textbf{\underline{Few-Shot PromptSmooth}}:  In \texttt{Few-Shot PromptSmooth}, we consider a scenario where only a few samples from the downstream medical task are available. Given a zero-shot classifier $f$ based on a pre-trained Med-VLM $(\mathbf{E}_{\text{image}},\mathbf{E}_{\text{text}})$ as well as a few labeled samples $\{(\mathbf{I}_{n}, y_{n})\}_{n=1}^N$ from a downstream dataset $\mathcal{D}$, where $\mathbf{I}_n \in \mathcal{I}$ and $y_n \in \mathcal{Y}$, \texttt{Few-Shot PromptSmooth} learns the prompts $\mathcal{P}$ as follows:
\begin{align}
\label{eq:p-tuning}
    \mathcal{P}^{\ast} = \argmin_{\mathcal{P}}\mathbb{E}_{\delta \sim \mathcal{N}(0, \sigma^2 \mathbf{I})} \frac{1}{N} \sum_{n=1}^{N}\mathcal{L}(f_{\mathcal{P}}(\mathbf{I}_n+ \delta), y_n), \\ 
    \nonumber
\end{align}
\noindent where $\mathcal{L}$ denotes the loss function between the classifier prediction and the ground-truth label. Similar to~\cite{zhou2022learning}, fine-tuning is performed to minimize the standard classification loss based on cross-entropy, and the gradients are back-propagated through the frozen text encoder $\mathbf{E}_{\mathtt{text}}$ to iteratively update the prompts $\mathcal{P}$. Note that these prompts are external to the pre-trained Med-VLM and they adjust the input context of the model without distorting its pre-trained features. Thus, this approach preserves the rich knowledge encoded in the frozen Med-VLMs to maintain high clean accuracy. 
At the same time, updating the prompts based on a few noisy samples from the downstream data set enhances certified robustness.\\
\textbf{\underline{Zero-Shot PromptSmooth}:}  In \texttt{Zero-Shot PromptSmooth}, the challenge is to learn the prompts $\mathcal{P}$ at inference time given only a single test sample $\mathbf{I}_t$ without any label. Given the lack of labels, the prompts cannot be optimized using the cross-entropy loss as in the few shot case. Therefore, we need a carefully designed unsupervised loss function for $\mathcal{L}$. Inspired by~\cite{shu2022test}, we optimize the prompts using a single step gradient descent based on the following entropy minimization loss.

\begin{align}
\label{eq:ent-min}
    \mathcal{P}^{\ast} = \argmin_{\mathcal{P}} \mathcal{H}\left(\mathbb{E}_{\delta \sim \mathcal{N}(0, \sigma^2 \mathbf{I})} \mathbb{Q}_{\mathcal{P}}(\mathbf{y}|(\mathbf{I}_t+\delta))\right),
\end{align}

\noindent where $\mathcal{H}$ denotes the entropy of a discrete probability distribution $\mathbb{Q}$, $\mathbb{Q}_{\mathcal{P}}(\mathbf{y}|(\mathbf{I}_t+\delta)) = [\mathbb{P}_{\mathcal{P}}(y_1|(\mathbf{I}_t+\delta)),\mathbb{P}_{\mathcal{P}}(y_2|(\mathbf{I}_t+\delta)),\cdots,\mathbb{P}_{\mathcal{P}}(y_K|(\mathbf{I}_t+\delta))]$, and $\mathbb{P}_{\mathcal{P}}(y_i|(\mathbf{I}_t+\delta))$ is the softmax output of the classifier $f_{\mathcal{P}}$ for class $y_i$, $i \in [1,K]$ based on the noisy input $(\mathbf{I}_t+\delta)$. Note that $\sum_{i=1}^{K}\mathbb{P}_{\mathcal{P}}(y_i|(\mathbf{I}_t+\delta)) = 1$. The above entropy minimization loss forces the classifier $f_{\mathcal{P}}$ to produce highly-confident (low entropy) yet consistent predictions for different noisy perturbations of $\mathbf{I}_t$. 

\noindent In practice, the expectation in both equations (\ref{eq:p-tuning}) and (\ref{eq:ent-min}) can be replaced by a sample average over $T$ Monte Carlo samples of $\delta$ drawn from Gaussian distributions with different values of $\sigma$ chosen from a desired range. This greatly reduces the computational cost of our approach because it avoids the need to learn the prompts $\mathcal{P}$ that are specific to a given $\sigma$. Finally, it is also possible to apply \texttt{Zero-Shot PromptSmooth} on top of \texttt{Few-Shot PromptSmooth} (i.e., combine both methods), which we simply refer to as \texttt{PromptSmooth}.

\section{Experiments} \label{sec:experiments}
\textbf{\underline{Models and Datasets}:} 
We evaluate our approach using three publicly available pre-trained Med-VLMs on six downstream datasets. The evaluated VLMs are PLIP~\cite{huang2023visual}, Quilt~\cite{ikezogwo2024quilt}, and MedCLIP~\cite{wang2022medclip}, with PLIP and Quilt trained on histopathology datasets and MedCLIP on X-ray images. Specifically, for PLIP, we show results on four pathology datasets: KatherColon~\cite{kather2019predicting}  (nine classes), PanNuke~\cite{gamper2019pannuke} (binary), SkinCancer~\cite{kriegsmann2022deep} (sixteen classes) and SICAPv2~\cite{silva2020going} (three classes). Quilt is evaluated on SkinCancer and SICAPv2, while MedCLIP is evaluated on the binary COVID~\cite{tawsifur2021zughaier}  and RSNA Pneumonia datasets~\cite{shih2019augmenting} (three-classes). For all of our experiments, we utilize the official train and test splits of the datasets unless otherwise mentioned.

\input{tables/certify_plip_kather_removecoop}
\input{tables/medclip_rsna_covid} 
\noindent \textbf{\underline{Implementation details}:} 
Our method is implemented in PyTorch on an NVIDIA A100 GPU with 40GB of memory. We report results with images normalized to $[0,1]^{224\times224\times3}$, aligning with prior studies. 
Labels for downstream dataset fine-tuning are converted into sentences, e.g., the label "Tumor" becomes `An H\&E image patch of $\{$Tumor$\}$'. 
For \texttt{Few-Shot PromptSmooth}, we fine-tune using a 16-shot setting for 50 epochs.
To update the prompts, we use SGD optimizer with a learning rate of 0.002 and a batch size of 16 and initialize prompts with 5 randomly initialized context tokens~\cite{zhou2022learning}. For \texttt{Zero-Shot PromptSmooth}, we augment with $T=100$ noisy samples and update the prompt with a single gradient descent step. For RS, we 
use $M = 10,000$ Monte Carlo samples with $\alpha = 0.001$ (see ~\cite{cohen2019certified}).\\
\noindent \textbf{\underline{Baselines}:} 
We conduct a comparative analysis of \texttt{PromptSmooth} against two representative RS techniques: \textit{Denoised Smoothing} \cite{salman2020denoised} and \textit{Diffusion Smoothing} \cite{carlini2022certified}, with the latter being the current state-of-the-art. Additionally, we also compare with zero-shot certification (no PL) and naive PL baselines. In the former scenario (see Sec.~\ref{sec:zeroshot}), certification results are obtained using hand-crafted prompts without PL, while naive PL~\cite{zhou2022learning} (CoOp) updates prompts using only clean samples from the target dataset.\\
\noindent \textbf{\underline{Evaluation}:} 
We use both clean and certified accuracy as the evaluation metrics. Certified accuracy is calculated as the proportion of the test set that \textsc{Certify}~\cite{cohen2019certified} correctly identifies for radius $R$ without abstention. Following prior works, we employ 
RS across four noise levels, $\sigma \in \{0.1, 0.25, 0.5, 1.0\}$, selecting the optimal results for each $R$ from 500 samples randomly chosen from the official test sets. 


\subsection{Results and Discussion}
\begin{figure}[t]
    \centering
    \begin{subfigure}[b]{0.32\textwidth}
        \includegraphics[width=\linewidth]{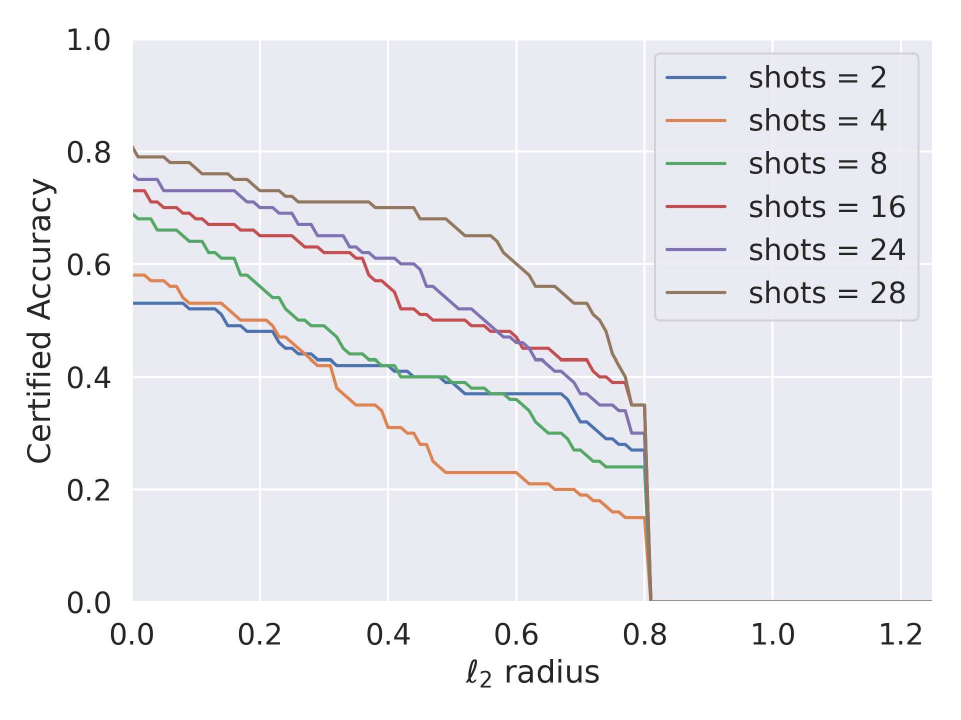}
        \caption{\small Shots}
        \label{fig:n_shots}
    \end{subfigure}
    \begin{subfigure}[b]{0.32\textwidth}
        \includegraphics[width=\linewidth]{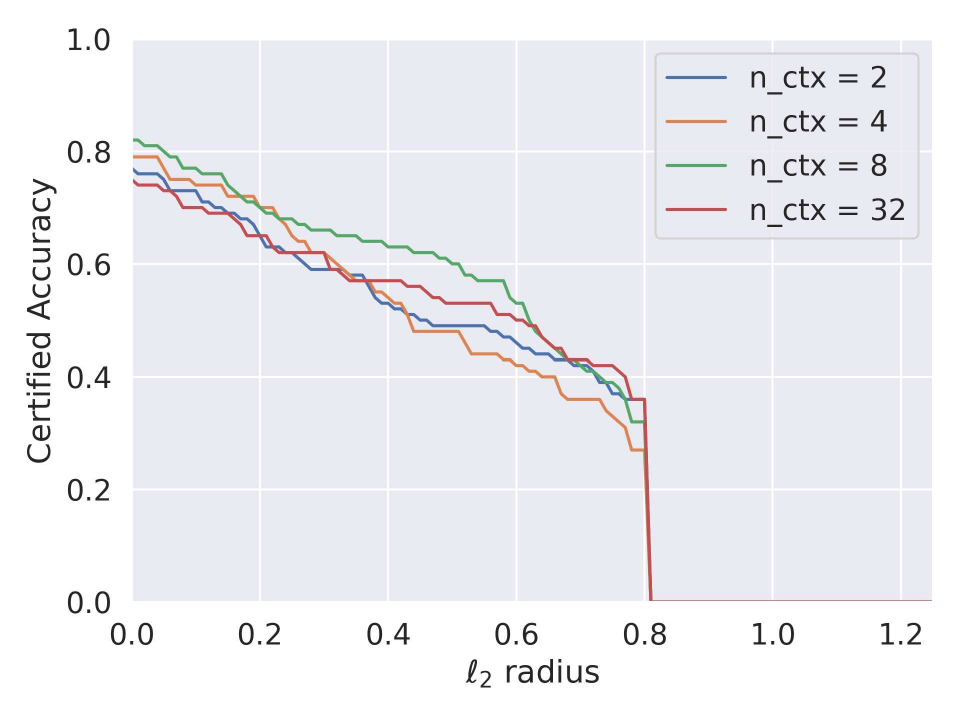}
        \caption{\small Context Tokens}
        \label{fig:n_ctx}
    \end{subfigure}
    \begin{subfigure}[b]{0.32\textwidth}
        \includegraphics[width=\linewidth]{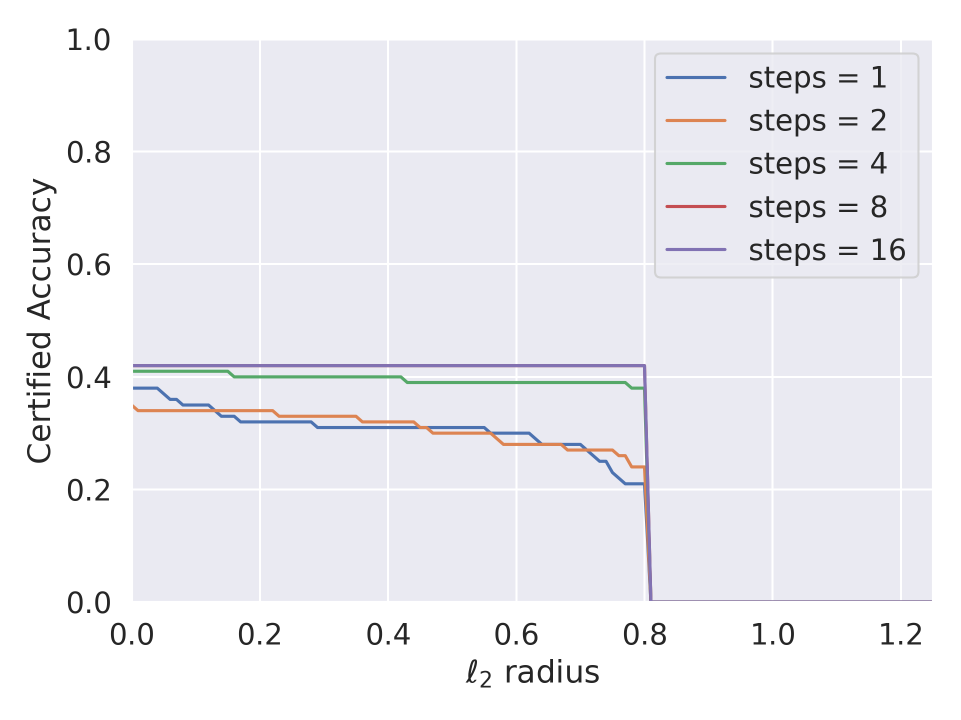}
        \caption{\small Optimizer Steps}
        \label{fig:tta_steps}
    \end{subfigure}
    \vspace{-0.1cm}
    \caption{\small Impact of changing the number of (a) shots and (b) context tokens in \texttt{Few-Shot PromptSmooth} and (c) varying the optimizer steps in \texttt{Zero-Shot PromptSmooth}.}
    \vspace{-0.5cm}
    \label{fig:ablation_plots}
\end{figure}
\input{tables/ctx_init_and_time}

Tab.~\ref{tab:plip_kather_results} compares \texttt{PromptSmooth} against baseline methods on the KatherColon dataset using the PLIP model. \texttt{PromptSmooth} consistently surpasses all baselines across each radius for both standard and certified accuracy. Notably, at a high radius of 1.5, it achieves an absolute gain of $29.6\%$ in certified accuracy over the recent \textit{Diffusion Smoothing} method.
\texttt{Zero-Shot PromptSmooth} outperforms baselines at higher radii by adapting to certifying input noise levels, while \texttt{Few-Shot PromptSmooth} achieves high certified accuracy at lower radii through alignment with noisy sample distributions via few-shot prompt learning. Combining zero-shot's adaptability with few-shot's noisy distribution alignment, \texttt{PromptSmooth} ensures high certified accuracy across all radii and maintains clean accuracy. Similar performance trends are observed in Tab.~\ref{tab:medclip_covid_rsna} for MedCLIP on the COVID and RSNA Pneumonia datasets.
Certification results for  Quilt and other datasets are provided in Appendix, and they show a similar trend.

\subsection{Ablations:}
All ablations are performed on the samples from the official test set of KatherColon dataset with PLIP model.

\noindent {\underline{Ablations for Few-Shot PromptSmooth}}: Increasing the number of samples per class in the few-shot case improves certified accuracy as depicted in Fig. \ref{fig:n_shots}, at the cost of a slight increase in fine-tuning time.  Additionally, Fig. \ref{fig:n_ctx} demonstrates that optimal certified accuracy is reached with 8 context tokens during PL, beyond which there is a degradation.

\noindent {\underline{Ablations for Zero-Shot PromptSmooth}}: Fig. \ref{fig:tta_steps} illustrates that certified accuracy increases with the number of gradient descent steps (up to 8), after which it plateaus. Additionally, initializing with a \texttt{noisy} context, as demonstrated in the Tab. \ref{tbl:ctx_init_zspromptsmooth}, enhances certified accuracy compared to standard prompts. 

\noindent {\underline{Computational Time}} As illustrated in Tab.~\ref{tbl:certification_time}, due to its lightweight nature, \texttt{PromptSmooth} is an order of magnitude faster than the Denoised Smoothing~\cite{salman2020denoised} and Diffusion Smoothing~\cite{carlini2022certified}.  Denoised Smoothing requires extensive training for custom denoisers, and Diffusion Smoothing which, despite utilizing pre-trained model, incurs longer certification time.

\section{Conclusion} In this paper, we introduced a novel approach for efficiently adapting a zero-shot classifier based on a Medical Vision-Language Model (Med-VLM) for adversarial robustness certification through prompt learning. We also developed two variants of our approach, specifically tailored for zero-shot and few-shot scenarios, which are particularly useful in the context of data-scarce medical applications. Extensive experiments conducted on three publicly available Med-VLMs and six downstream datasets demonstrate that our proposed approach achieves state-of-the-art performance. Moreover, it is computationally efficient and does not require large medical datasets, which enhances its practicality. 

\begin{credits}
\subsubsection{\discintname}
\noindent The authors have no competing interests to declare that are relevant to the content of this article. 
\end{credits}

\bibliographystyle{splncs04}
\bibliography{refs}

\end{document}


\appendix

\noindent\begin{LARGE} \textbf{Appendix} \vspace{4mm} \end{LARGE}

\begin{table}[h!]
\centering
\caption{\small Certification results of PLIP on 3 datasets for zero-shot PLIP, naive prompt learning and our proposed \texttt{PromptSmooth} methods.}
\resizebox{\linewidth}{!}{
\begin{tabular}{lccccccc}
\toprule
\rowcolor{gray!20} \textbf{Method} & \multicolumn{7}{c}{\textbf{Certified Accuracy at $\ell_2$ radius (\%)}} \\
\rowcolor{gray!20} & 0.1 & 0.25 & 0.5 & 0.75 & 1.0 & 1.25 & 1.5 \\
\midrule
\multicolumn{8}{c}{\textbf{PanNuke (Binary)}} \\
\midrule
Zero-shot PLIP (No PL) & $^{(59.6)}$55.8 & $^{(52)}$50.8 & $^{(52)}$50.2 & $^{(50.2)}$50.0 & $^{(50.2)}$50.0 & $^{(50.2)}$50.0 & $^{(50.2)}$50.0 \\
Naive PL (CoOp) \cite{zhou2022learning} & $^{(73.6)}$69.8 & $^{(73.6)}$61.8 & $^{(50.0)}$50.0 & $^{(50.0)}$50.0 & $^{(50.0)}$50.0 & $^{(50.0)}$50.0 & $^{(50.0)}$50.0 \\
\rowcolor{Apricot!20} Zero-shot PromptSmooth & $^{(51.6)}$51.4 & $^{(51.6)}$51.4 & $^{(50.6)}$50.6 & $^{(50.6)}$50.6 & $^{(50.6)}$50.6 & $^{(50.6)}$50.6 & $^{(50.6)}$50.6 \\
\rowcolor{Apricot!20} Few-shot PromptSmooth & $^{(73.2)}$68.8 & $^{(73.2)}$58.8 & $^{(50.0)}$50.0 & $^{(50.0)}$50.0 & $^{(50.0)}$50.0 & $^{(50.0)}$50.0 & $^{(50.0)}$50.0 \\
\rowcolor{Apricot!20} PromptSmooth & $^{(73.8)}$\textbf{73.8 }& $^{(73.8)}$\textbf{73.8 }& $^{(68.8)}$\textbf{68.8} & $^{(68.8)}$\textbf{68.8} & $^{(65.2)}$\textbf{65.2} & $^{(65.2)}$\textbf{65.2} & $^{(65.2)}$\textbf{65.2} \\
\midrule
\multicolumn{8}{c}{\textbf{SICAPv2 (4 Classes)}} \\
\midrule
Zero-shot PLIP (No PL) & $^{(43.2)}$40.0 & $^{(43.2)}$33.6 & $^{(25.0)}$25.5 & $^{(25.0)}$25.0 & $^{(25.0)}$25.0 & $^{(25.0)}$25.0 & $^{(25.0)}$25.0 \\
Naive PL (CoOp) \cite{zhou2022learning} & $^{(45.0)}$37.0 & $^{(45.0)}$27.2 & $^{(25.0)}$25.5 & $^{(25.0)}$25.0 & $^{(25.0)}$25.0 & $^{(25.0)}$25.0 & $^{(25.0)}$25.0 \\
\rowcolor{Apricot!20} Zero-shot PromptSmooth & $^{(44.2)}$42.8 & $^{(44.2)}$39.6 & $^{(44.2)}$33.8 & $^{(44.2)}$27.8 & $^{(25.0)}$25.0 & $^{(25.0)}$25.0 & $^{(25.0)}$25.0 \\
\rowcolor{Apricot!20} Few-shot PromptSmooth & $^{(51.4)}$40.0 & $^{(51.4)}$26.6 & $^{(27.8)}$15.8 & $^{(27.8)}$10.8 & $^{(27.8)}$7.40 & $^{(27.8)}$4.40 & $^{(27.8)}$2.00 \\
\rowcolor{Apricot!20} PromptSmooth & $^{(51.6)}$\textbf{50.8} & $^{(51.6)}$\textbf{49.6} & $^{(46.8)}$\textbf{39.4} & $^{(46.8)}$\textbf{35.0} & $^{(46.8)}$\textbf{30.2} & $^{(46.8)}$\textbf{30.0} & $^{(46.8)}$\textbf{28.2} \\
\midrule
\multicolumn{8}{c}{\textbf{SkinCancer (16 Classes)}} \\
\midrule
Zero-shot PLIP (No PL) & $^{(26.6)}$19.0 & $^{(14.2)}$9.80 & $^{(6.0)}$5.80 & $^{(6.00)}$4.80 & $^{(6.00)}$4.60 & $^{(6.00)}$4.39 & $^{(6.00)}$4.39 \\
Naive PL (CoOp) \cite{zhou2022learning} & $^{(67.2)}$57.8 & $^{(67.2)}$43.8 & $^{(34.2)}$16.6 & $^{(34.2)}$7.00 & $^{(6.20)}$6.20 & $^{(6.20)}$6.20 & $^{(6.20)}$6.20 \\
\rowcolor{Apricot!20} Zero-shot PromptSmooth & $^{(25)}$23.2 & $^{(25)}$19.6 & $^{(14.2)}$11.6 & $^{(6.40)}$6.40 & $^{(6.40)}$6.40 & $^{(6.40)}$6.40 & $^{(6.40)}$6.40 \\
\rowcolor{Apricot!20} Few-shot PromptSmooth & $^{(67.0)}$58.0 & $^{(67.0)}$44.6 & $^{(54.4)}$27.0 & $^{(36.6)}$15.0 & $^{(36.6)}$9.00 & $^{(6.20)}$6.20 & $^{(6.20)}$6.20 \\
\rowcolor{Apricot!20} PromptSmooth & $^{(68.6)}$\textbf{67.6} & $^{(68.6)}$\textbf{64.8} & $^{(56.4)}$\textbf{50.2} & $^{(56.4)}$\textbf{46.0} & $^{(38.2)}$\textbf{33.8} & $^{(38.2)}$\textbf{31.0}& $^{(38.2)}$\textbf{29.6} \\
\bottomrule
\end{tabular}
}
\label{tab:combined-results}
\end{table}

\begin{table}[h!]
\centering
\caption{\small Certification results for Quilt on 2 datasets.}
\resizebox{\linewidth}{!}{
\begin{tabular}{lccccccc}
\toprule
\rowcolor{gray!20} \textbf{Method} & \multicolumn{7}{c}{\textbf{Certified Accuracy at $\ell_2$ radius (\%)}} \\
\rowcolor{gray!20} & 0.1 & 0.25 & 0.5 & 0.75 & 1 & 1.25 & 1.5 \\
\midrule
\multicolumn{8}{c}{\textbf{SICAPv2 (4 Classes)}}\\
\midrule
Zero-shot Quilt (No PL) & $^{(25.0)}$25.0 & $^{(25.0)}$25.0 & $^{(25.0)}$25.0 & $^{(25)}$25.0 & $^{(25)}$24.6 & $^{(25)}$24.2 & $^{(25)}$23.8 \\
Naive PL (CoOp) \cite{zhou2022learning} & $^{(55.8)}$47.2 & $^{(55.8)}$35.8 & $^{(25.0)}$25.0 & $^{(25.0)}$25.0 & $^{(25.0)}$25.0 & $^{(25.0)}$25.0 & $^{(25.0)}$25.0 \\
\rowcolor{Apricot!20} Zero-shot Promptsmooth & $^{(25.8)}$25.8 & $^{(25.8)}$25.6 & $^{(25.8)}$25.6 & $^{(25.8)}$25.2 & $^{(25.8)}$25.0 & $^{(25.8)}$24.8 & $^{(25.8)}$24.8 \\
\rowcolor{Apricot!20} Few-shot PromptSmooth & $^{(62.0)}$49.8 & $^{(62.0)}$31.8 & $^{(31.2)}$17.0 & $^{(31.2)}$10.8 & $^{(31.2)}$6.40 & $^{(31.2)}$2.60 & $^{(31.2)}$0.60 \\
\rowcolor{Apricot!20} PromptSmooth & $^{(60.0)}$\textbf{59.0} & $^{(60.0)}$\textbf{57.0} & $^{(55.4)}$\textbf{49.8} & $^{(55.4)}$\textbf{46.8} & $^{(32.2)}$\textbf{31.0} & $^{(32.2)}$\textbf{30.6} & $^{(32.2)}$\textbf{30.6} \\
\midrule
\multicolumn{8}{c}{\textbf{SkinCancer (16 Classes)}} \\
\midrule
Zero-shot Quilt (No PL) & $^{(33.0)}$25.8 & $^{(33.0)}$15.8 & $^{(6.20)}$6.20 & $^{(6.20)}$6.20 & $^{(6.20)}$6.20 & $^{(6.20)}$6.20 & $^{(6.20)}$6.20 \\
Naive PL (CoOp) \cite{zhou2022learning} & $^{(52.8)}$43.2 & $^{(52.8)}$31.0 & $^{(18.2)}$7.60 & $^{(6.80)}$6.20 & $^{(6.80)}$6.20 & $^{(6.80)}$6.20 & $^{(6.80)}$6.20 \\
\rowcolor{Apricot!20} Zero-shot PromptSmooth & $^{(29.2)}$26.4 & $^{(29.2)}$20.6 & $^{(14.79)}$9.80 & $^{(14.79)}$6.60 & $^{(4.20)}$3.20 & $^{(4.20)}$3.20 & $^{(4.20)}$2.40 \\
\rowcolor{Apricot!20} Few-shot PromptSmooth & $^{(69.4)}$61.0 & $^{(64.4)}$51.0 & $^{(64.4)}$53.0 & $^{(64.4)}$35.0 & $^{(64.4)}$17.6 & $^{(46.0)}$9.20 & $^{(46.0)}$3.80 \\
\rowcolor{Apricot!20} PromptSmooth & $^{(71.0)}$\textbf{70.0} & $^{(71.0)}$\textbf{67.0} & $^{(66.2)}$\textbf{62.0} & $^{(66.2)}$\textbf{57.8} & $^{(46.8)}$\textbf{42.6} & $^{(46.8)}$\textbf{41.6} & $^{(46.8)}$\textbf{40.6} \\
\bottomrule
\end{tabular}
}
\label{tab:quilt_datasets}
\end{table}

\bibliographystyle{splncs04}
\bibliography{appendix_refs}

%% file: tables/certify_plip_kather_removecoop.tex
\begin{table}[t]
\centering
\caption{\small Certification results for PLIP on KatherColon dataset, where the numbers indicate certified accuracy (\%). Corresponding clean accuracy (\%) is in parentheses.}
\resizebox{\linewidth}{!}{
\begin{tabular}{lccccccc}
\toprule
\rowcolor{gray!20} 
\textbf{Method} & \multicolumn{7}{c}{\textbf{Certified Accuracy at $\ell_2$ radius (\%)}} \\
\rowcolor{gray!20} & 0.1 & 0.25 & 0.5 & 0.75 & 1.0 & 1.25 & 1.5\\
\midrule
Zero-shot PLIP (No PL) & $^{(56.6)}$49.4 & $^{(56.6)}$38.2 & $^{(28.9)}$20.8 & $^{(28.9)}$17.6 & $^{(11.0)}$11.0 & $^{(11.0)}$11.0 & $^{(11.0)}$11.0 \\
Naive PL (CoOp) \cite{zhou2022learning} & $^{(71.6)}$66.7 & $^{(71.6)}$56.0 & $^{(22.0)}$16.4 & $^{(22.0)}$14.2 & $^{(11.0)}$11.0 & $^{(11.0)}$11.0 & $^{(11.0)}$11.0 \\
Denoised Smoothing \cite{salman2020denoised} & $^{(55.0)}$48.2 & $^{(55.0)}$39.2 & $^{(45.2)}$31.0 & $^{(45.2)}$25.6 & $^{(26.2)}$17.4 & $^{(26.2)}$16.2 & $^{(26.2)}$14.6 \\
Diffusion Smoothing \cite{carlini2022certified} & $^{(58.0)}$57.0 & $^{(53.0)}$49.0 & $^{(53.0)}$41.0	& $^{(53.0)}$34.0 & $^{(53.0)}$26.0 & $^{(53.0)}$22.0	& $^{(53.0)}$16.0 \\
\rowcolor{Apricot!20} \texttt{Zero-shot PromptSmooth} & $^{(57.6)}$53.4 & $^{(57.6)}$49.0 & $^{(30.2)}$29.0 & $^{(30.2)}$29.0 & $^{(30.2)}$28.6 & $^{(30.2)}$28.4 & $^{(30.2)}$27.4 \\
\rowcolor{Apricot!20}  \texttt{Few-Shot PromptSmooth} & $^{(81.2)}$78.2 & $^{(81.2)}$67.6 & $^{(75.6)}$52.2 & $^{(75.6)}$35.6 & $^{(50.4)}$26.4 & $^{(50.4)}$22.2 & $^{(50.4)}$17.6 \\
\midrule
\rowcolor{Apricot!20} \texttt{PromptSmooth}& $^{(82.0)}$\textbf{81.8} & $^{(82.0)}$\textbf{81.0} & $^{(76.6)}$\textbf{74.8} & $^{(76.6)}$\textbf{73.2} & $^{(54.0)}$\textbf{48.4} & $^{(54.0)}$\textbf{47.2} & $^{(54.0)}$\textbf{45.6} \\
\bottomrule
\end{tabular}
}
\label{tab:plip_kather_results}
\end{table}

%% file: tables/medclip_rsna_covid.tex
\begin{table}[t]
\centering
\caption{\small Certified accuracy (\%) for MedCLIP on COVID and RSNA Pneumonia datasets, with the corresponding clean accuracy (\%) in parentheses.}
\resizebox{\linewidth}{!}{
\begin{tabular}{lccccc|cccc}
\toprule
\rowcolor{gray!20} \textbf{Method} & \multicolumn{4}{c}{\textbf{COVID}} & & \multicolumn{4}{c}{\textbf{RSNA Pneumonia}} \\
\rowcolor{gray!20}& 0.1 & 0.25 & 0.5 & 0.75 & & 0.1 & 0.25 & 0.5 & 0.75 \\
\midrule
Denoised Smoothing \cite{salman2020denoised} & $^{(66.4)}$54.6 & $^{(50.2)}$48.4 & $^{(50.2)}$45.8 & $^{(50.2)}$36.0 & & $^{(37.6)}$27.6 & $^{(31.6)}$21.0 & $^{(31.6)}$9.40 & $^{(31.6)}$1.79 \\
Diffusion Smoothing \cite{carlini2022certified} & $^{(56.0)}$37.0 & $^{(44.0)}$22.0 & $^{(44.0)}$6.00 & $^{(44.0)}$1.00 & & $^{(44.0)}$40.0 & $^{(44.0)}$28.0 & $^{(44.0)}$12.0 & $^{(44.0)}$1.00 \\
\rowcolor{Apricot!20} \texttt{Zero-shot PromptSmooth} & $^{(62.4)}$62.0 & $^{(62.4)}$60.6 & $^{(50.2)}$50.0 & $^{(50.2)}$49.8 & & $^{(37.0)}$35.4 & $^{(37.0)}$33.4 & $^{(33.4)}$33.4 & $^{(33.4)}$\textbf{33.2} \\
\rowcolor{Apricot!20} \texttt{Few-shot PromptSmooth} & $^{(66.8)}$58.0 & $^{(52.0)}$48.8 & $^{(52.0)}$47.6 & $^{(52)}$42.8 & & $^{(41.4)}$34.4 & $^{(34.0)}$31.2 & $^{(34.0)}$27.0 & $^{(34.0)}$23.6 \\
\midrule
\rowcolor{Apricot!20} \texttt{PromptSmooth} & $^{(69.4)}$\textbf{69.0} & $^{(69.4)}$\textbf{68.4} & $^{(53.0)}$\textbf{52.8} & $^{(53.0)}$\textbf{52.6} & & $^{(42.4)}$\textbf{40.8} & $^{(42.4)}$\textbf{35.8} & $^{(34.6)}$\textbf{33.4} & $^{(34.6)}$32.0 \\
\bottomrule
\end{tabular}
}
\label{tab:medclip_covid_rsna}
\end{table}

%% file: tables/ctx_init_and_time.tex
\begin{table}[t]
\centering
\begin{minipage}{0.475\textwidth}
\caption{\small \texttt{Zero-shot PromptSmooth} context initialization}
\label{tbl:ctx_init_zspromptsmooth}
\resizebox{\linewidth}{!}{\begin{tabular}{lcccc}
\toprule
\rowcolor{gray!20} \textbf{Prompt} & 
 \multicolumn{4}{c} {\textbf{$\ell_2$ radius}} \\
 \rowcolor{gray!20} & 0 & 0.1 & 0.25 & 0.5\\
\midrule
``An H\&E image patch of'' & 38.0 & 35.0 & 32.0& 31.0 \\
``An H\&E noisy image patch of'' & \textbf{40.0} & \textbf{38.0} & \textbf{38.0} & \textbf{35.0} \\
\bottomrule
\end{tabular}}
\end{minipage}
\begin{minipage}{0.465\textwidth}
\caption{\small{Training time and certification time per sample}}
\label{tbl:certification_time}
\resizebox{\linewidth}{!}{
\begin{tabular}{lc|c|c}
\toprule
\rowcolor{gray!20}\textbf{Method} & \multicolumn{3}{c}{\textbf{Time}} \\
\rowcolor{gray!20} & Training & Certification &  Total\\
\midrule
Denoised Smoothing~\cite{salman2020denoised} & 8h 47m &  17s & 8h 47m 17s \\
Diffusion Smoothing~\cite{carlini2022certified} & - & 4m 40s & 4m 40s \\
\texttt{PromptSmooth} & 40s & 1.9s & 41.9s\\
\bottomrule
\end{tabular}}
\end{minipage}
\end{table}

%% file: main.bbl
\begin{thebibliography}{10}
\providecommand{\url}[1]{\texttt{#1}}
\providecommand{\urlprefix}{URL }
\providecommand{\doi}[1]{https://doi.org/#1}

\bibitem{athalye2018obfuscated}
Athalye, A., Carlini, N., Wagner, D.: Obfuscated gradients give a false sense of security: Circumventing defenses to adversarial examples. In: International conference on machine learning. pp. 274--283. PMLR (2018)

\bibitem{azad2023foundational}
Azad, B., Azad, R., Eskandari, S., Bozorgpour, A., Kazerouni, A., Rekik, I., Merhof, D.: Foundational models in medical imaging: A comprehensive survey and future vision. arXiv preprint arXiv:2310.18689  (2023)

\bibitem{carlini2022certified}
Carlini, N., Tramer, F., Dvijotham, K.D., Rice, L., Sun, M., Kolter, J.Z.: (certified!!) adversarial robustness for free! arXiv preprint arXiv:2206.10550  (2022)

\bibitem{cohen2019certified}
Cohen, J., Rosenfeld, E., Kolter, Z.: Certified adversarial robustness via randomized smoothing. In: international conference on machine learning. pp. 1310--1320. PMLR (2019)

\bibitem{dong2023adversarial}
Dong, J., Chen, J., Xie, X., Lai, J., Chen, H.: Adversarial attack and defense for medical image analysis: Methods and applications. arXiv preprint arXiv:2303.14133  (2023)

\bibitem{finlayson2019adversarial}
Finlayson, S.G., Bowers, J.D., Ito, J., Zittrain, J.L., Beam, A.L., Kohane, I.S.: Adversarial attacks on medical machine learning. Science  \textbf{363}(6433),  1287--1289 (2019)

\bibitem{gamper2019pannuke}
Gamper, J., Alemi~Koohbanani, N., Benet, K., Khuram, A., Rajpoot, N.: Pannuke: an open pan-cancer histology dataset for nuclei instance segmentation and classification. In: Digital Pathology: 15th European Congress, ECDP 2019, Warwick, UK, April 10--13, 2019, Proceedings 15. pp. 11--19. Springer (2019)

\bibitem{han2023medical}
Han, T., Nebelung, S., Khader, F., Wang, T., Mueller-Franzes, C., F{\"o}rsch, S., Kleesiek, C., Bressem, K.K., et~al.: Medical foundation models are susceptible to targeted misinformation attacks. arXiv preprint arXiv:2309.17007  (2023)

\bibitem{huang2023visual}
Huang, Z., Bianchi, F., Yuksekgonul, M., Montine, T.J., Zou, J.: A visual--language foundation model for pathology image analysis using medical twitter. Nature medicine  \textbf{29}(9),  2307--2316 (2023)

\bibitem{ikezogwo2024quilt}
Ikezogwo, W., Seyfioglu, S., Ghezloo, F., Geva, D., Sheikh~Mohammed, F., Anand, P.K., Krishna, R., Shapiro, L.: Quilt-1m: One million image-text pairs for histopathology. Advances in Neural Information Processing Systems  \textbf{36} (2024)

\bibitem{kather2019predicting}
Kather, J.N., Krisam, J., Charoentong, P., Luedde, T., Herpel, E., Weis, C.A., Gaiser, T., Marx, A., Valous, N.A., Ferber, D., et~al.: Predicting survival from colorectal cancer histology slides using deep learning: A retrospective multicenter study. PLoS medicine  \textbf{16}(1),  e1002730 (2019)

\bibitem{kriegsmann2022deep}
Kriegsmann, K., Lobers, F., Zgorzelski, C., Kriegsmann, J., Janssen, C., Meliss, R.R., Muley, T., Sack, U., Steinbuss, G., Kriegsmann, M.: Deep learning for the detection of anatomical tissue structures and neoplasms of the skin on scanned histopathological tissue sections. Frontiers in Oncology  \textbf{12},  1022967 (2022)

\bibitem{kumari2023trust}
Kumari, A., Bhardwaj, D., Jindal, S., Gupta, S.: Trust, but verify: A survey of randomized smoothing techniques. arXiv preprint arXiv:2312.12608  (2023)

\bibitem{laousy2023certification}
Laousy, O., Araujo, A., Chassagnon, G., Paragios, N., Revel, M.P., Vakalopoulou, M.: Certification of deep learning models for medical image segmentation. In: International Conference on Medical Image Computing and Computer-Assisted Intervention. pp. 611--621. Springer (2023)

\bibitem{lecuyer2019certified}
Lecuyer, M., Atlidakis, V., Geambasu, R., Hsu, D., Jana, S.: Certified robustness to adversarial examples with differential privacy. In: 2019 IEEE symposium on security and privacy (SP). pp. 656--672. IEEE (2019)

\bibitem{li2023sok}
Li, L., Xie, T., Li, B.: Sok: Certified robustness for deep neural networks. In: 2023 IEEE symposium on security and privacy (SP). pp. 1289--1310. IEEE (2023)

\bibitem{qiu2023exploring}
Qiu, K., Zhang, H., Wu, Z., Lin, S.: Exploring transferability for randomized smoothing. arXiv preprint arXiv:2312.09020  (2023)

\bibitem{radford2021learning}
Radford, A., Kim, J.W., Hallacy, C., Ramesh, A., Goh, G., Agarwal, S., Sastry, G., Askell, A., Mishkin, P., Clark, J., et~al.: Learning transferable visual models from natural language supervision. In: International conference on machine learning. pp. 8748--8763. PMLR (2021)

\bibitem{salman2020denoised}
Salman, H., Sun, M., Yang, G., Kapoor, A., Kolter, J.Z.: Denoised smoothing: A provable defense for pretrained classifiers. Advances in Neural Information Processing Systems  \textbf{33},  21945--21957 (2020)

\bibitem{shih2019augmenting}
Shih, G., Wu, C.C., Halabi, S.S., Kohli, M.D., Prevedello, L.M., Cook, T.S., Sharma, A., Amorosa, J.K., Arteaga, V., Galperin-Aizenberg, M., et~al.: Augmenting the national institutes of health chest radiograph dataset with expert annotations of possible pneumonia. Radiology: Artificial Intelligence  \textbf{1}(1),  e180041 (2019)

\bibitem{shrestha2023medical}
Shrestha, P., Amgain, S., Khanal, B., Linte, C.A., Bhattarai, B.: Medical vision language pretraining: A survey. arXiv preprint arXiv:2312.06224  (2023)

\bibitem{shu2022test}
Shu, M., Nie, W., Huang, D.A., Yu, Z., Goldstein, T., Anandkumar, A., Xiao, C.: Test-time prompt tuning for zero-shot generalization in vision-language models. Advances in Neural Information Processing Systems  \textbf{35},  14274--14289 (2022)

\bibitem{silva2023foundation}
Silva-Rodriguez, J., Chakor, H., Kobbi, R., Dolz, J., Ayed, I.B.: A foundation language-image model of the retina (flair): Encoding expert knowledge in text supervision. arXiv preprint arXiv:2308.07898  (2023)

\bibitem{silva2020going}
Silva-Rodr{\'\i}guez, J., Colomer, A., Sales, M.A., Molina, R., Naranjo, V.: Going deeper through the gleason scoring scale: An automatic end-to-end system for histology prostate grading and cribriform pattern detection. Computer methods and programs in biomedicine  \textbf{195},  105637 (2020)

\bibitem{tawsifur2021zughaier}
Tawsifur, R., Amith, K., Yazan, Q., Anas, T., Serkan, K., Abul, K.S.B., Tariqul, I.M., Somaya, A.M.: Zughaier susu m, khan muhammad salman, et al. Exploring the effect of image enhancement techniques on covid-19 detection using chest x-ray images. Computers in biology and medicine  \textbf{132},  104319 (2021)

\bibitem{wang2022medclip}
Wang, Z., Wu, Z., Agarwal, D., Sun, J.: Medclip: Contrastive learning from unpaired medical images and text. arXiv preprint arXiv:2210.10163  (2022)

\bibitem{zhang2023prompt}
Zhang, J., Kapse, S., Ma, K., Prasanna, P., Saltz, J., Vakalopoulou, M., Samaras, D.: Prompt-mil: Boosting multi-instance learning schemes via task-specific prompt tuning. arXiv preprint arXiv:2303.12214  (2023)

\bibitem{zhao2024evaluating}
Zhao, Y., Pang, T., Du, C., Yang, X., Li, C., Cheung, N.M.M., Lin, M.: On evaluating adversarial robustness of large vision-language models. Advances in Neural Information Processing Systems  \textbf{36} (2024)

\bibitem{zhao2023clip}
Zhao, Z., Liu, Y., Wu, H., Li, Y., Wang, S., Teng, L., Liu, D., Li, X., Cui, Z., Wang, Q., et~al.: Clip in medical imaging: A comprehensive survey. arXiv preprint arXiv:2312.07353  (2023)

\bibitem{zhong2023ariadne}
Zhong, Y., Xu, M., Liang, K., Chen, K., Wu, M.: Ariadne’s thread: Using text prompts to improve segmentation of infected areas from chest x-ray images. In: International Conference on Medical Image Computing and Computer-Assisted Intervention. pp. 724--733. Springer (2023)

\bibitem{zhou2022learning}
Zhou, K., Yang, J., Loy, C.C., Liu, Z.: Learning to prompt for vision-language models. International Journal of Computer Vision  \textbf{130}(9),  2337--2348 (2022)

\end{thebibliography}
